\documentclass[11pt]{article}

\usepackage[final]{acl}

\usepackage{times}
\usepackage{latexsym}
\usepackage{amsmath}
\usepackage{amsfonts}
\usepackage{amssymb}
\usepackage[T1]{fontenc}
\usepackage[utf8]{inputenc}
\usepackage{booktabs}
\usepackage{multirow}
\usepackage{microtype}
\usepackage{graphicx}
\usepackage{xcolor}
\usepackage{url}
\usepackage{hyperref}
\usepackage{wrapfig}
\usepackage{enumitem}
\usepackage{tikz}
\usepackage{placeins}

\usetikzlibrary{positioning, arrows.meta, fit, backgrounds}
\usepackage{CJKutf8}

\definecolor{lightred}{RGB}{255, 245, 245}
\definecolor{lightblue}{RGB}{245, 250, 255}
\definecolor{darkblue}{RGB}{0, 0, 139}
\definecolor{taggray}{RGB}{100, 100, 100}
\definecolor{softred}{RGB}{190, 30, 30}
\definecolor{arrowcolor}{RGB}{80, 80, 80}
\definecolor{processblue}{RGB}{0, 50, 150}
\definecolor{customblue}{RGB}{65, 105, 225}

\title{

MMed-Bench-IR: A Heterogeneous Benchmark for Multilingual Medical Information Retrieval

}

\hypersetup{
    colorlinks=true,
    linkcolor=darkblue,
    filecolor=magenta,
    urlcolor=processblue,
    citecolor=darkblue,
}

\author{
  \textbf{Junhyeok Lee\textsuperscript{1,2,$\dagger$}},
  \textbf{Han Jang\textsuperscript{1,3,$\dagger$}},
  \textbf{Hyeonjin Goh\textsuperscript{3}},
  \textbf{Kyu Sung Choi\textsuperscript{1,2,3,$*$}}
\\
  \small \textsuperscript{1}Seoul National University \quad
  \textsuperscript{2}Seoul National University College of Medicine \quad
  \textsuperscript{3}Seoul National University Hospital \\
  \small \textsuperscript{$\dagger$}Equal contribution \quad
  \textsuperscript{$*$}Corresponding author \\
  \small \texttt{ent1127@snu.ac.kr}\\
}

\begin{document}
\maketitle

\begin{abstract}

Retrieval-augmented generation~(RAG) in clinical settings increasingly requires multilingual retrieval against predominantly English evidence corpora.
Multilingual medical retrieval demands three capabilities: cross-lingual alignment, concept discrimination, and evidence retrieval.
However, existing benchmarks evaluate these only in isolation, leaving the interaction between biomedical expertise and multilingual coverage unmeasured.
We introduce \textbf{MMed-Bench-IR}, a benchmark designed to disentangle these axes across 6~languages and three structurally heterogeneous tasks:
(1)~cross-lingual medical QA retrieval with 6{,}127 queries grounded in the Unified Medical Language System~(UMLS),
(2)~concept discrimination over 4{,}975 confusion sets at three difficulty tiers,
and (3)~multilingual evidence retrieval for RAG with 2{,}040 quality-assured queries.
The three tasks share zero concept and query overlap by design, ensuring that aggregate scores reflect genuine capability breadth.
Evaluation of ten systems across six paradigm families reveals severe cross-lingual failure: biomedical encoders that score 0.818~nDCG@10 in English drop to 0.056 in Japanese, a gap that English-only benchmarks cannot detect.

\end{abstract}

\section{Introduction}
\label{sec:intro}

\begin{figure}[t]
  \centering
  \includegraphics[width=\linewidth]{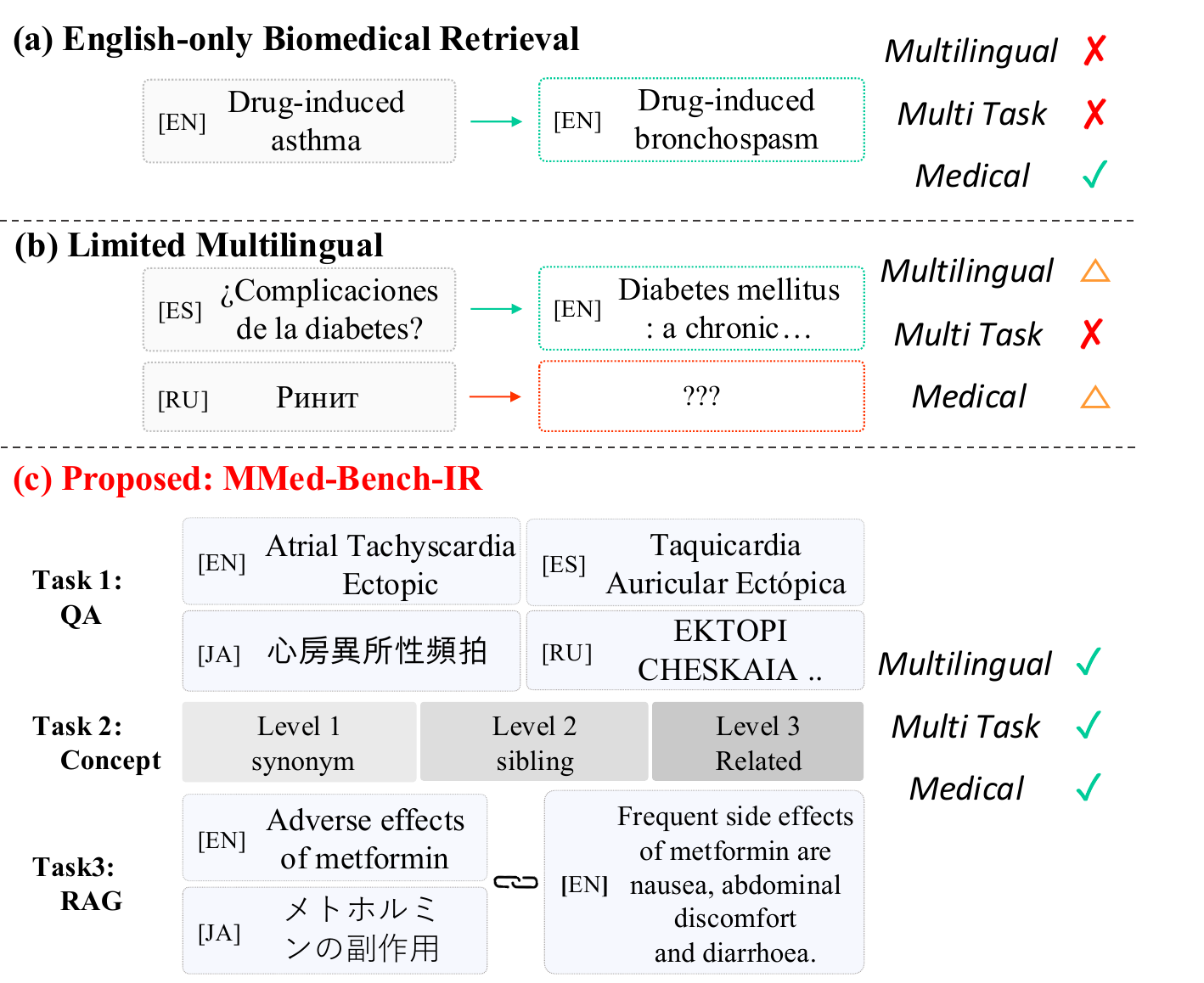}
    \caption{\textbf{Motivation.}
    (a)~English-only biomedical benchmarks miss non-English queries.
    (b)~Existing multilingual medical benchmarks cover 2--3 languages and one task.
    (c)~MMed-Bench-IR evaluates all three axes across 6~languages and 3~writing systems.}
  \label{fig:motivation}
\end{figure}

Large language models are increasingly adopted in healthcare~\citep{sahni2023artificial}, and the quality of their clinical responses depends heavily on the retrieval systems that supply supporting evidence~\citep{qiu2024towards,wang2024apollo}.
However, these retrieval systems are predominantly built for and evaluated in English, creating a growing gap for non-English-speaking populations.
Language barriers are associated with reduced access to care, higher rates of adverse events, and poorer health outcomes across clinical settings~\citep{joo2023association}, and recent work warns that AI-powered clinical tools risk deepening these inequities by exhibiting significant performance disparities between well-resourced and digitally under-represented languages~\citep{anyaegbuna2026artificial,ortega2025language}.

A key bottleneck lies in the retrieval stage: embedding models and dense retrievers that power these systems often lack the multilingual alignment needed to handle diverse languages~\citep{yuan2022coder,liu2021self}, yet the field has no way to measure this shortcoming systematically.
This multilingual medical retrieval problem decomposes into three distinct capabilities: cross-lingual alignment so that the same concept in different languages maps to similar representations, concept discrimination so that clinically confusable entities such as Type~1 versus Type~2 diabetes are correctly separated, and evidence retrieval so that queries in any language can surface relevant English-language passages.
These three capabilities are individually well studied~\citep{liu2021self,remy2023biolord,zhang2023miracl}, yet no existing benchmark evaluates them jointly~\citep{thakur2021beir,muennighoff2023mteb}, leaving the field unable to measure whether progress on one dimension comes at the cost of another.

Prior work addresses each axis only partially:
biomedical encoders such as SapBERT~\citep{liu2021self} and BioLORD-2023~\citep{remy2023biolord} evaluate only in English, while multilingual benchmarks such as MIRACL~\citep{zhang2023miracl} lack medical concept grounding.
Recent efforts in cross-lingual medical retrieval~\citep{athar2025cure,acharya-etal-2025-m3retrieve} cover limited language pairs or target multimodal rather than text-only retrieval.
The intersection of multilingual coverage and biomedical specialization remains uncharted (Figure~\ref{fig:motivation}; Section~\ref{sec:related}).

To address these gaps, we introduce \textbf{MMed-Bench-IR}, a structured evaluation suite spanning 6~languages (English, Spanish, French, Japanese, Chinese, Russian) and 3~writing systems (Latin, Cyrillic, CJK).
Inspired by the Benchmarking Information Retrieval~(BEIR) suite~\citep{thakur2021beir}, which showed that heterogeneous evaluation across diverse tasks and domains yields insights invisible in any single benchmark, we design MMed-Bench-IR around the same principle applied to multilingual medical information retrieval.

It comprises three structurally heterogeneous tasks: cross-lingual medical question answering~(QA) retrieval with 6{,}127 queries grounded in the Unified Medical Language System~(UMLS) Metathesaurus ontology~\cite{bodenreider2004unified}, concept discrimination over 4{,}975 confusion sets organized into three difficulty levels, and multilingual evidence retrieval for retrieval-augmented generation~(RAG) with 2{,}040 quality-assured queries against an 80{,}049-passage English corpus.
By construction, the three tasks share zero annotated concept overlap, zero query overlap, and minimal corpus vocabulary overlap, ensuring that aggregate scores reflect genuine capability breadth rather than mastery of a single skill.

Our contributions are as follows:
\begin{itemize}
  \item To the best of our knowledge, MMed-Bench-IR is the first benchmark to jointly evaluate multilingual alignment and biomedical specialization across 6~languages, 3~writing systems, and three tasks with zero concept and query overlap.
  \item We evaluate ten systems across six paradigm families and reveal a consistent paradigm-level hierarchy across all tasks, with concept discrimination emerging as the hardest lexical bottleneck at a +0.38 gap between BM25 and the best dense model.
  \item We expose severe cross-lingual failure modes invisible in English-only evaluation: biomedical encoders that score 0.818~nDCG@10 in English collapse to 0.056 in Japanese, with a fairness gap of 0.76.
\end{itemize}

\begin{figure*}[!ht]
\centering
\includegraphics[width=\linewidth]{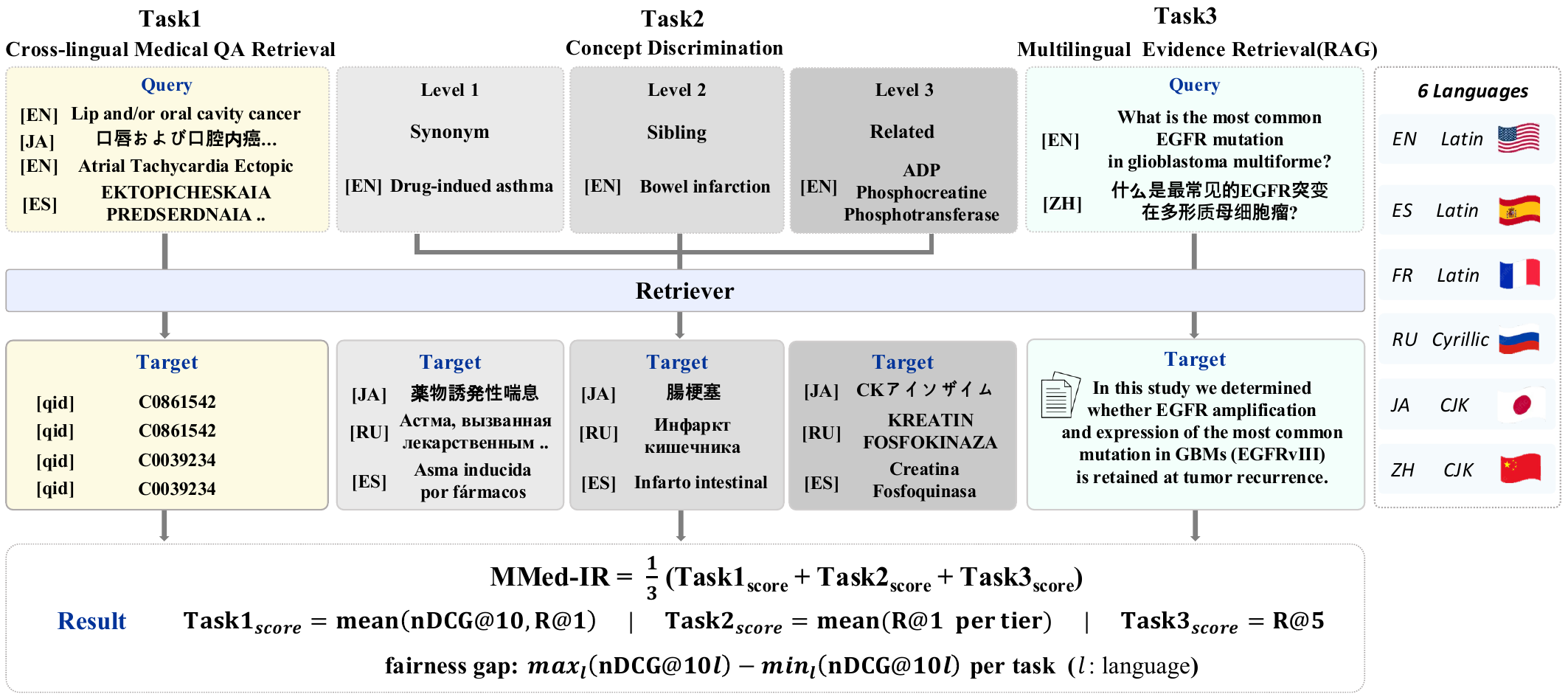}
\caption{\textbf{Overview of MMed-Bench-IR}. Three tasks target distinct retrieval capabilities across 6~languages with zero query and CUI overlap. Scores aggregate into MMed-IR and per-task fairness gaps.}
\label{fig:overview}
\end{figure*}

\section{Related Work and Scope}
\label{sec:related}

\paragraph{Multilingual and biomedical retrieval models.}
Recent work has made significant progress on multilingual and biomedical retrieval independently, but the two axes remain largely disconnected.
MMedC~\citep{qiu2024towards} and Apollo~\citep{wang2024apollo} curate multilingual medical corpora for LLM training but provide no retrieval evaluation.
On the retriever side, BGE-M3~\citep{chen-etal-2024-m3} and Multilingual-E5~\citep{wang2024multilingual} support over 100~languages but lack medical concept grounding, while BioLORD-2023~\citep{remy2023biolord} and MedCPT~\citep{jin2023medcpt} advance biomedical entity representation but evaluate only in English.
UMLS-based~\citep{bodenreider2004unified} entity linking work such as SapBERT~\citep{liu2021self} and CODER~\citep{yuan2022coder} tests concept alignment but not passage retrieval.
No existing model jointly addresses both axes.

\paragraph{Retrieval benchmarks.}
Several benchmarks have advanced retrieval evaluation along individual axes, but none jointly targets multilingual and biomedical capabilities.
BEIR~\citep{thakur2021beir} established that in-domain performance does not predict out-of-domain generalization, motivating heterogeneous evaluation.
MTEB~\citep{muennighoff2023mteb} demonstrated that no single embedding method dominates across diverse tasks, reinforcing the need for multi-task assessment.
However, neither targets the biomedical domain.
On the multilingual side, MIRACL~\citep{zhang2023miracl} and Mr.~TyDi~\citep{zhang2021mr} cover diverse languages but do not incorporate medical terminology or ontology structure.
Despite these advances, no existing benchmark can evaluate the interaction between multilingual coverage and biomedical specialization.

\paragraph{Cross-lingual medical retrieval.}
A small but growing body of work has begun to address cross-lingual retrieval in the medical domain directly.
The MUCHMORE project~\citep{volk2002semantic} explored UMLS-based cross-language information retrieval~(CLIR) over English and German medical abstracts, establishing medical CLIR as a research direction.
More recently, CURE~\citep{athar2025cure} evaluates point-of-care passage ranking across English, French-to-English, and Spanish-to-English conditions but covers only three languages and a single retrieval task.
M3Retrieve~\citep{acharya-etal-2025-m3retrieve} targets multimodal medical retrieval rather than multilingual text-only information retrieval.
MMed-Bench-IR differs by jointly evaluating concept alignment, concept discrimination, and evidence retrieval within a single benchmark spanning 6~languages and 3~writing systems.

\section{Benchmark Design Principles}
\label{sec:design}

We design MMed-Bench-IR around the principle that benchmark value comes from principled heterogeneity rather than dataset count, and organize it along four selection criteria.

\paragraph{Capability heterogeneity.}
The three tasks isolate distinct retrieval capabilities: cross-lingual concept alignment (Task~1), fine-grained semantic discrimination (Task~2), and cross-lingual passage retrieval (Task~3).
This separation ensures that a model's aggregate score cannot be driven by strength in a single capability.

\paragraph{Linguistic heterogeneity.}
Six languages span three language families (Indo-European [en, es, fr, ru], Sino-Tibetan [zh], Japonic [ja]), three writing systems (Latin, Cyrillic, CJK), and varying levels of medical NLP resource availability.
This ensures that benchmark scores reflect genuine multilingual robustness rather than performance on closely related languages.

\paragraph{Difficulty heterogeneity.}
Task~1 spans 500 medical concepts across 5~languages with varying UMLS coverage, creating natural difficulty gradients across languages.
Task~2 includes three difficulty tiers (synonym, sibling, related-but-distinct) validated by multi-encoder consensus.
Task~3 uses quality-assured translations with an 83.7\% retention rate.
The resulting score distributions span a 5$\times$ range across systems, with no task saturated.

\paragraph{Validation heterogeneity.}
Each task uses a different validation strategy targeting a different error source: UMLS ontology grounding (Task~1), multi-encoder majority vote (Task~2), and bilingual concept-fidelity and back-translation QA (Task~3).
This ensures that no single validation blind spot propagates across tasks.

\begin{table*}[!ht]
\centering
\caption{MMed-Bench-IR benchmark statistics.
The three tasks differ structurally in retrieval setting, corpus scale, query granularity, and relevance definition, ensuring that aggregate scores reflect capability breadth.}
\label{tab:stats}
\small
\begin{tabular}{@{}p{2.8cm}ccc@{}}
\toprule
& \textbf{Task 1: QA} & \textbf{Task 2: Concept} & \textbf{Task 3: RAG} \\
\midrule
Retrieval setting & Q$_\text{lang}$ $\to$ C$_\text{multi}$ & Q$_\text{term}$ $\to$ C$_\text{pool}$ & Q$_\text{multi}$ $\to$ C$_\text{en}$ \\
Query languages & 5 & 2 (en, zh) & 6 \\
Corpus language & Multilingual & Multilingual & English \\
\# Queries & 6{,}127 & 4{,}143 & 2{,}040 \\
\# Corpus docs & 2{,}552 & 6{,}340 & 80{,}049 \\
Avg.\ query len.\ (words) & 5.0 & 4.2 & 6.7 \\
Avg.\ doc len.\ (words) & 5.7 & 5.2 & 30.1 \\
Avg.\ pos.\ per query & 1.9 & 1.5 & 32.9 \\
Relevance definition & CUI alignment & Same-CUI synonym & BioASQ gold \\
Validation method & UMLS grounding & 3-encoder vote & Concept fidelity + BT \\
Primary metric & mean(nDCG, R@1) & mean(R@1 per tier) & R@5 \\
\bottomrule
\end{tabular}
\end{table*}

\section{Benchmark Construction and Characterization}
\label{sec:benchmark}

\subsection{Benchmark Statistics}

Table~\ref{tab:stats} and Figure~\ref{fig:overview} summarize the benchmark.
The three tasks are structurally complementary: they share zero CUI overlap between Task~1 and Task~2 concept spaces, and the pairwise corpus vocabulary overlap is low (Jaccard $<$0.14 between Tasks~1/2, $<$0.02 between either and Task~3).
Query granularity ranges from short concept terms (4 to 5 words) to biomedical questions (6.7 words), and corpus scale spans two orders of magnitude (2{,}552 to 80{,}049 documents).

\subsection{Task Construction}

\paragraph{Task~1: Cross-lingual Medical QA Retrieval.}
From MMedBench~\cite{qiu2024towards} test-split multilingual medical QA data, we extract question stems (without answer options) as queries across 5~languages (French is excluded because fewer than 50 queries passed the CUI-tagging pipeline at this granularity).
Each query is tagged with a UMLS CUI via a cascading tagger that first attempts exact lexical match against UMLS preferred terms and synonyms, then falls back to a biomedical linker for unmatched queries.
Queries sharing the same CUI across languages form cross-lingual positive groups.
The corpus consists of concept definition texts (1{,}290 positives + 1{,}262 hard-negative concepts from non-matching CUIs), yielding 6{,}127 queries and 2{,}552 corpus documents across 500 concepts.

\paragraph{Task~2: Concept Discrimination.}
From the UMLS 2025AB Metathesaurus concept file (MRCONSO.RRF)~\citep{bodenreider2004unified}, we parse 1.5M concept atoms grouped by source vocabulary and construct confusion sets at three tiers:
Tier~1 (synonym) contains the same CUI with different surface forms or languages.
Tier~2 (sibling) pairs different CUIs sharing the same source vocabulary and 3-character code prefix, meaning they are clinically related but distinct (e.g., ICD-10 codes under the same category).
Tier~3 (related) pairs different CUIs from the same source vocabulary but different code prefix groups, meaning they belong to the same medical domain but are clearly distinct.
Queries are drawn from English and Chinese surface forms, and positives are different surface forms of the same CUI in any language, with the query's own form excluded from the corpus.
All 6{,}340 corpus terms across multiple languages form a shared retrieval pool.
Because each confusion set generates one query per unique CUI it contains, the 4{,}975 validated sets yield 4{,}143 evaluation queries.

\paragraph{Task~3: Multilingual Evidence Retrieval for RAG.}
340 English biomedical questions from BioASQ~13b~\citep{tsatsaronis2015overview} with 80{,}049 evidence snippets are translated to 5~target languages using NLLB-200-3.3B~\citep{costa2022no}.
Translation quality is assured via two checks.
The first is concept fidelity, where each translation is verified against a 587K-entry UMLS bilingual lexicon plus a set of 87 protected medical abbreviations (e.g., HIV, MRI, HbA1c) that must be preserved.
The second is back-translation consistency, where each translation is back-translated via NLLB-200 and compared with the original using token-overlap F1.
Queries passing at least one check form the official evaluation subset (1{,}708 of 2{,}040, 83.7\%).

\subsection{Heterogeneity Analysis}

\begin{figure}[t]
\centering
\includegraphics[width=\linewidth]{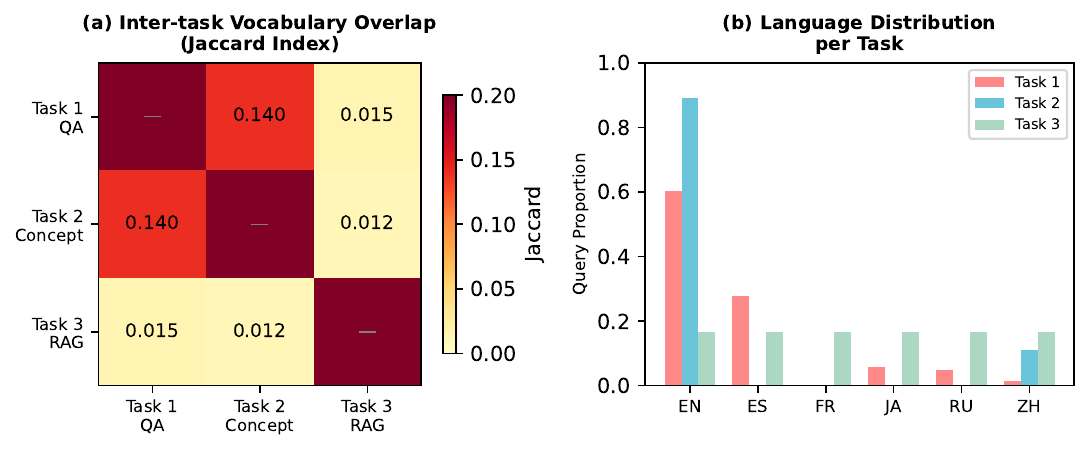}
\caption{\textbf{Benchmark heterogeneity.}
(a)~Inter-task corpus vocabulary overlap is low (Jaccard $\leq$0.14), confirming that the three tasks target distinct lexical domains.
(b)~Language distributions vary across tasks: Task~1 is English-dominant due to UMLS coverage, Task~2 uses English and Chinese only, and Task~3 is uniformly balanced across all six languages.}
\label{fig:heterogeneity}
\end{figure}

Figure~\ref{fig:heterogeneity} confirms that the three tasks are structurally independent.
Vocabulary overlap between Task~3 (biomedical passages) and either term-level task is below 0.02 Jaccard, reflecting fundamentally different retrieval regimes.
The language distributions differ by design: Task~1 reflects natural UMLS coverage (English-dominant), Task~2 captures the two most distinct scripts (en, zh), and Task~3 is uniformly balanced.
Zero queries overlap between any two tasks, and zero annotated CUIs overlap between Tasks~1 and~2 (Task~3 uses BioASQ questions without CUI annotation).
These properties ensure that a model's aggregate performance cannot be inflated by excelling on a single lexical domain or language.

\subsection{Validation}

Benchmark validity rests on three controls, each targeting a different error source.

\paragraph{Task~1.}
Ontology-grounded positives via UMLS CUI alignment ensure face validity.

\paragraph{Task~2.}
Tier definitions are anchored in UMLS ontology structure (synonym, sibling, related-but-distinct), providing an encoder-independent criterion.
A 3-encoder majority vote using BGE-M3~\citep{chen-etal-2024-m3}, E5-large~\citep{wang2024multilingual}, and SapBERT~\citep{liu2021self} operationalizes acceptance thresholds based on pairwise cosine similarity (Tier~1: $\geq$0.70, Tier~2: 0.40--0.75, Tier~3: 0.15--0.45), requiring agreement from at least two of three encoders.
This process rejects 967 sets (16.3\%) and re-tiers 1{,}597, yielding 4{,}975 validated sets.
To verify that the tier ordering is not biased toward the validators, we confirm that BM25 and BioLORD-2023~\citep{remy2023biolord}, neither of which participated in validation, exhibit the same monotonic Tier~1, Tier~2, Tier~3 ordering (Table~\ref{tab:tiers}), with tier boundaries confirmed by non-overlapping confidence intervals (Appendix~\ref{app:construction}).

\paragraph{Task~3.}
Two-stage translation quality assurance based on concept fidelity and back-translation consistency retains 83.7\% of queries.
Full thresholds and per-language breakdowns are in Appendix~\ref{app:construction}.

\subsection{Scoring}

The primary score \textbf{MMed-IR} is the macro average of three task scores:
\begin{equation}
    \text{MMed-IR} = \frac{1}{3}\bigl(\text{Task1}_\text{score} + \text{Task2}_\text{score} + \text{Task3}_\text{score}\bigr)
\end{equation}
where Task~1 score $=$ mean(nDCG@10, R@1), Task~2 score $=$ mean(R@1$_\text{Tier1}$, R@1$_\text{Tier2}$, R@1$_\text{Tier3}$), and Task~3 score $=$ R@5.
The fairness gap for each task is $\max_\ell(\text{nDCG@10}_\ell) - \min_\ell(\text{nDCG@10}_\ell)$ across languages~$\ell$, providing a uniform cross-lingual equity measure.

\begin{table*}[!ht]
\centering
\caption{MMed-Bench-IR main results with query-level bootstrap 95\% confidence intervals ($n{=}2{,}000$).
All paradigm-level differences exceed CI widths. Best in \textbf{bold}, second-best \underline{underlined}. $^\dagger$Within-distribution.}
\label{tab:main}
\small
\begin{tabular}{@{}lccccc@{}}
\toprule
\textbf{Model} & \textbf{Task 1} & \textbf{Task 2} & \textbf{Task 3} & \textbf{MMed-IR} & \textbf{Family} \\
\midrule
\multicolumn{6}{@{}l}{\textit{Zero-shot}} \\
BM25 & .091\,{\scriptsize$\pm$.005} & .076\,{\scriptsize$\pm$.006} & .056\,{\scriptsize$\pm$.008} & .075\,{\scriptsize$\pm$.004} & Lexical \\
BioLORD & .188\,{\scriptsize$\pm$.006} & .213\,{\scriptsize$\pm$.012} & .088\,{\scriptsize$\pm$.008} & .163\,{\scriptsize$\pm$.005} & Bio-dense \\
SapBERT & .212\,{\scriptsize$\pm$.006} & .282\,{\scriptsize$\pm$.010} & .107\,{\scriptsize$\pm$.009} & .201\,{\scriptsize$\pm$.005} & Bio-dense \\
BM25+BGE-M3 & .173\,{\scriptsize$\pm$.005} & .124\,{\scriptsize$\pm$.007} & .191\,{\scriptsize$\pm$.012} & .163\,{\scriptsize$\pm$.005} & Hybrid \\
ColBERT-XM & .268\,{\scriptsize$\pm$.007} & .372\,{\scriptsize$\pm$.009} & .244\,{\scriptsize$\pm$.013} & .295\,{\scriptsize$\pm$.006} & Late-int \\
BGE-M3 & .344\,{\scriptsize$\pm$.007} & .432\,{\scriptsize$\pm$.010} & .264\,{\scriptsize$\pm$.013} & .347\,{\scriptsize$\pm$.006} & Multi-dense \\
E5-large & .335\,{\scriptsize$\pm$.008} & \underline{.447}\,{\scriptsize$\pm$.012} & .262\,{\scriptsize$\pm$.013} & .348\,{\scriptsize$\pm$.007} & Multi-dense \\
BGE-M3+Reranker & .387\,{\scriptsize$\pm$.007} & .433\,{\scriptsize$\pm$.010} & \underline{.294}\,{\scriptsize$\pm$.014} & .371\,{\scriptsize$\pm$.006} & Two-stage \\
\midrule
\multicolumn{6}{@{}l}{\textit{Within-distribution}$^\dagger$} \\
MMed-Embed & .388\,{\scriptsize$\pm$.007} & \underline{.454}\,{\scriptsize$\pm$.011} & .267\,{\scriptsize$\pm$.013} & .369\,{\scriptsize$\pm$.006} & Multi+Med \\
MMed-Embed+Rer. & \textbf{.404}\,{\scriptsize$\pm$.007} & .436\,{\scriptsize$\pm$.010} & .291\,{\scriptsize$\pm$.014} & \textbf{.377}\,{\scriptsize$\pm$.006} & Multi+Med+R \\
\bottomrule
\end{tabular}
\end{table*}

\section{Experimental Setup}
\label{sec:experiments}

\subsection{Baselines}

We evaluate ten systems spanning six paradigm families to characterize the benchmark's discriminative properties:
\begin{itemize}
    \item \textbf{Lexical}: BM25~\citep{robertson2009probabilistic} (Okapi, whitespace tokenization; no CJK-specific segmentation, which disadvantages BM25 on Japanese/Chinese).
    \item \textbf{Biomedical dense}: SapBERT~\citep{liu2021self} and BioLORD-2023~\citep{remy2023biolord}, English-centric biomedical encoders.
    \item \textbf{Multilingual dense}: BGE-M3~\citep{chen-etal-2024-m3} and E5-large~\citep{wang2024multilingual}, general-purpose encoders supporting over 100~languages.
    \item \textbf{Late-interaction}: ColBERT-XM~\citep{louis2025colbert}, a multilingual late-interaction retriever based on X-MOD~\citep{pfeiffer-etal-2022-lifting} with per-token scoring, trained on MS-MARCO.
    \item \textbf{Hybrid}: BM25~\citep{robertson2009probabilistic} and BGE-M3~\citep{chen-etal-2024-m3} combined via reciprocal rank fusion (RRF, $k{=}60$).
    \item \textbf{Two-stage reranker}: BGE-M3~\citep{chen-etal-2024-m3} retrieval followed by BGE-reranker-v2-m3 cross-encoder reranking (top-30).
    \item \textbf{Multilingual and medical}: MMed-Embed (Section~\ref{sec:reference}) and MMed-Embed with Reranker, within-distribution references.
\end{itemize}
MMed-Embed combines both biomedical specialization and multilingual coverage, providing an approximate within-distribution ceiling.

\subsection{Reference Model: MMed-Embed}
\label{sec:reference}

To assess the benchmark's discriminative range above existing baselines, we fine-tune BGE-M3~\citep{chen-etal-2024-m3} (568M parameters, XLM-RoBERTa~\citep{conneau2020unsupervised} backbone) on 137K training rows from MMedBench~\citep{qiu2024towards} (40K, 6~languages), UMLS~\citep{bodenreider2004unified} cross-lingual pairs (87K, 5~languages), and xBioASQ (10K).
No overlap exists with benchmark test sets (verified: 0\% query overlap on Task~3; $<$0.5\% trivial term overlap on Tasks~1/2).

\paragraph{Training objective.}
$\mathcal{L}_\text{ret}$ is temperature-scaled cross-entropy over cosine similarities between each query and all in-batch positives plus $k{=}15$ hard negatives, with temperature $\tau{=}0.02$.

\paragraph{Training procedure.}
We fine-tune BGE-M3 on the full 137K rows with 15 random in-batch negatives per query for 10~epochs (effective batch 1{,}024, lr $2{\times}10^{-6}$, 1{,}330~steps) using AdamW with linear warmup (10\%) and linear decay, mean pooling with L2 normalization, max sequence length 256, gradient checkpointing, and mixed precision (bf16) on 8$\times$B200~GPUs.
A 2$\times$2 ablation (Appendix~\ref{app:reference}) shows that neither topology-aware margin loss nor ANCE-style hard negative mining~\citep{xiong2020approximate} yields statistically significant gains over this baseline recipe; the primary benefit comes from domain-specific contrastive fine-tuning on multilingual medical data.
Full hyperparameters are in Appendix~\ref{app:reference}.

\paragraph{Results.}
MMed-Embed achieves 0.369 MMed-IR, a marginal +0.022 over the zero-shot BGE-M3 baseline (0.347).
Combining MMed-Embed with cross-encoder reranking yields the best overall score (0.377) and the lowest fairness gap (0.170), showing that domain-specific first-stage retrieval and cross-encoder reranking are complementary.
The modest gain from 137K-row fine-tuning suggests that the benchmark's primary challenge lies in architecture-level multilingual alignment rather than in training data or loss design; MMed-Embed scores should be interpreted as an approximate within-distribution ceiling rather than a fair zero-shot comparison.

\section{Results and Analysis}
\label{sec:results}

\subsection{Overall Comparison}

Table~\ref{tab:main} presents the main results across all ten systems.
The spread from BM25 (0.075) to the best system (0.377) spans a 5$\times$ range, confirming effective discrimination across paradigm families, with substantial headroom remaining.

\subsection{Key Findings}

\paragraph{Finding A: Consistent hierarchy across paradigms and tasks.}
All three tasks produce the same paradigm ordering: lexical methods trail biomedical-only encoders, which in turn trail multilingual dense retrievers (Table~\ref{tab:main}).
This ordering holds across all six paradigm families, including ColBERT-XM (0.295), which falls between biomedical encoders (SapBERT 0.201) and multilingual dense retrievers (BGE-M3 0.347), consistent with its English-centric contrastive training.
Hybrid RRF (0.163) underperforms standalone dense retrieval, while cross-encoder reranking (0.371) and MMed-Embed with Reranker (0.377) occupy the top tier.
This consistency across tasks with zero query and CUI overlap cannot be attributed to data leakage.

\paragraph{Finding B: Concept discrimination is the hardest lexical bottleneck.}
Task~2 most sharply separates paradigms: BM25 scores 0.076 versus the best model at 0.454, a +0.38 gap (Table~\ref{tab:main}).
Hybrid RRF degrades Task~2 performance (0.124 vs.\ 0.432 for BGE-M3 alone), confirming that lexical signal is counterproductive for concept discrimination.
Tier~3 (related-but-distinct concepts) remains a genuine frontier: even the strongest model reaches only 3.6\%, and ColBERT-XM's token-level scoring yields only 0.8\% (Table~\ref{tab:tiers}).

\paragraph{Finding C: Biomedical specialization does not rescue cross-lingual failure.}
SapBERT achieves 0.818~nDCG@10 in English but collapses to 0.056 in Japanese (Table~\ref{tab:languages}), with a fairness gap of 0.76$\pm$0.04 versus 0.24$\pm$0.05 for BGE-M3, a 3$\times$ equity difference with non-overlapping CIs.
ColBERT-XM exhibits the same pattern: its Task~3 fairness gap (0.37) exceeds BGE-M3 (0.24), confirming that late-interaction scoring alone does not rescue cross-lingual retrieval when contrastive training is English-centric.
Cross-encoder reranking achieves the lowest fairness gap (0.18$\pm$0.04).
Full per-task gaps are in Appendix~\ref{app:full_results}.

\subsection{Evaluation Impact}

Each finding requires the multi-axis evaluation that MMed-Bench-IR provides: the consistent hierarchy emerges only from multi-task comparison, the lexical bottleneck only from concept-level retrieval, and the cross-lingual collapse only from joint multilingual and biomedical evaluation.

\begin{table}[t]
\centering
\caption{Task~2 Recall@1 by difficulty tier with query-level bootstrap 95\% CIs.}
\label{tab:tiers}
\small
\begin{tabular}{@{}lccc@{}}
\toprule
\textbf{Model} & \textbf{Tier~1} & \textbf{Tier~2} & \textbf{Tier~3} \\
& {\scriptsize(Synonym)} & {\scriptsize(Sibling)} & {\scriptsize(Related)} \\
\midrule
\multicolumn{4}{@{}l}{\textit{Zero-shot}} \\
BM25 & .172\,{\scriptsize$\pm$.014} & .053\,{\scriptsize$\pm$.009} & .004\,{\scriptsize$\pm$.006} \\
BioLORD & .449\,{\scriptsize$\pm$.018} & .166\,{\scriptsize$\pm$.017} & .025\,{\scriptsize$\pm$.025} \\
SapBERT & .606\,{\scriptsize$\pm$.018} & .233\,{\scriptsize$\pm$.020} & .008\,{\scriptsize$\pm$.010} \\
BM25+BGE-M3 & .281\,{\scriptsize$\pm$.016} & .087\,{\scriptsize$\pm$.013} & .004\,{\scriptsize$\pm$.006} \\
ColBERT-XM & .799\,{\scriptsize$\pm$.014} & .309\,{\scriptsize$\pm$.021} & .008\,{\scriptsize$\pm$.010} \\
BGE-M3 & \underline{.892}\,{\scriptsize$\pm$.010} & .384\,{\scriptsize$\pm$.022} & .019\,{\scriptsize$\pm$.017} \\
E5-large & \underline{.892}\,{\scriptsize$\pm$.010} & \underline{.419}\,{\scriptsize$\pm$.022} & \underline{.029}\,{\scriptsize$\pm$.028} \\
BGE-M3+Rer. & .862\,{\scriptsize$\pm$.011} & .406\,{\scriptsize$\pm$.020} & .031\,{\scriptsize$\pm$.021} \\
\midrule
\multicolumn{4}{@{}l}{\textit{Within-distribution}} \\
MMed-Embed & \textbf{.893}\,{\scriptsize$\pm$.010} & \textbf{.435}\,{\scriptsize$\pm$.021} & \textbf{.033}\,{\scriptsize$\pm$.022} \\
MMed+Rer. & .866\,{\scriptsize$\pm$.011} & .419\,{\scriptsize$\pm$.021} & .023\,{\scriptsize$\pm$.017} \\
\bottomrule
\end{tabular}
\end{table}

\begin{table*}[t]
\centering
\caption{Task~3 nDCG@10 by query language with bootstrap 95\% CIs and fairness gap. Gap CIs computed via bootstrap of max$-$min across languages.}
\label{tab:languages}
\small
\begin{tabular}{@{}lccccccc@{}}
\toprule
\textbf{Model} & \textbf{EN} & \textbf{ES} & \textbf{FR} & \textbf{JA} & \textbf{ZH} & \textbf{RU} & \textbf{Gap} \\
\midrule
\multicolumn{8}{@{}l}{\textit{Zero-shot}} \\
BM25 & .518\,{\scriptsize$\pm$.04} & .082\,{\scriptsize$\pm$.02} & .122\,{\scriptsize$\pm$.03} & .063\,{\scriptsize$\pm$.02} & .036\,{\scriptsize$\pm$.02} & .110\,{\scriptsize$\pm$.03} & .482\,{\scriptsize$\pm$.04} \\
BioLORD & .755\,{\scriptsize$\pm$.04} & .358\,{\scriptsize$\pm$.04} & .500\,{\scriptsize$\pm$.04} & .062\,{\scriptsize$\pm$.02} & .081\,{\scriptsize$\pm$.02} & .067\,{\scriptsize$\pm$.02} & .693\,{\scriptsize$\pm$.04} \\
SapBERT & .818\,{\scriptsize$\pm$.03} & .452\,{\scriptsize$\pm$.05} & .501\,{\scriptsize$\pm$.04} & .056\,{\scriptsize$\pm$.02} & .095\,{\scriptsize$\pm$.03} & .113\,{\scriptsize$\pm$.03} & .762\,{\scriptsize$\pm$.04} \\
BM25+BGE-M3 & .731\,{\scriptsize$\pm$.03} & .498\,{\scriptsize$\pm$.03} & .533\,{\scriptsize$\pm$.03} & .369\,{\scriptsize$\pm$.03} & .330\,{\scriptsize$\pm$.03} & .453\,{\scriptsize$\pm$.03} & .401\,{\scriptsize$\pm$.04} \\
ColBERT-XM & .870\,{\scriptsize$\pm$.02} & .795\,{\scriptsize$\pm$.03} & .757\,{\scriptsize$\pm$.03} & .527\,{\scriptsize$\pm$.05} & .502\,{\scriptsize$\pm$.04} & .747\,{\scriptsize$\pm$.04} & .369\,{\scriptsize$\pm$.05} \\
BGE-M3 & .855\,{\scriptsize$\pm$.03} & .836\,{\scriptsize$\pm$.03} & .823\,{\scriptsize$\pm$.03} & .679\,{\scriptsize$\pm$.04} & .620\,{\scriptsize$\pm$.04} & .798\,{\scriptsize$\pm$.03} & .235\,{\scriptsize$\pm$.05} \\
E5-large & \underline{.893}\,{\scriptsize$\pm$.02} & .840\,{\scriptsize$\pm$.03} & .827\,{\scriptsize$\pm$.03} & .659\,{\scriptsize$\pm$.04} & .499\,{\scriptsize$\pm$.04} & .814\,{\scriptsize$\pm$.03} & .394\,{\scriptsize$\pm$.05} \\
BGE-M3+Reranker & \underline{.895}\,{\scriptsize$\pm$.02} & \underline{.876}\,{\scriptsize$\pm$.02} & \underline{.876}\,{\scriptsize$\pm$.02} & \underline{.772}\,{\scriptsize$\pm$.04} & \underline{.715}\,{\scriptsize$\pm$.04} & \underline{.860}\,{\scriptsize$\pm$.03} & \underline{.180}\,{\scriptsize$\pm$.04} \\
\midrule
\multicolumn{8}{@{}l}{\textit{Within-distribution}} \\
MMed-Embed & .853\,{\scriptsize$\pm$.03} & .830\,{\scriptsize$\pm$.03} & .821\,{\scriptsize$\pm$.03} & .688\,{\scriptsize$\pm$.04} & .632\,{\scriptsize$\pm$.04} & .787\,{\scriptsize$\pm$.03} & .221\,{\scriptsize$\pm$.05} \\
MMed-Embed+Rer. & \textbf{.892}\,{\scriptsize$\pm$.02} & \textbf{.867}\,{\scriptsize$\pm$.03} & \textbf{.869}\,{\scriptsize$\pm$.03} & \textbf{.776}\,{\scriptsize$\pm$.04} & \textbf{.722}\,{\scriptsize$\pm$.04} & \textbf{.847}\,{\scriptsize$\pm$.03} & \textbf{.170}\,{\scriptsize$\pm$.04} \\
\bottomrule
\end{tabular}
\end{table*}

\section{Benchmark Validity and Bias Analysis}
\label{sec:validity}

\paragraph{Annotation bias.}
All three tasks use automated relevance judgments rather than human annotations, eliminating inter-annotator disagreement but introducing potential systematic biases.
For Task~2, tier definitions are anchored in encoder-independent UMLS ontology relations, and both BM25 and BioLORD-2023 (neither of which participated in validation) exhibit the same monotonic tier ordering (Table~\ref{tab:tiers}).
A leave-one-validator-out analysis confirms $\geq$85.4\% tier stability across all encoder-pair conditions (Appendix~\ref{app:construction}).

\paragraph{Translation bias.}
Task~3 uses NLLB-200 machine translation, which may introduce systematic quality differences across languages.
Model ranking is perfectly preserved between the official (1{,}708~queries) and all-translated (2{,}040) subsets (Spearman $\rho{=}1.0$), confirming that QA filtering does not introduce ranking artifacts (Appendix~\ref{app:construction}).

\paragraph{LLM-based quality audit.}
We conduct a 600-sample quality audit using Claude Opus 4.7~(Anthropic, 2026) across all 12 task$\times$language cells (50~per cell; Table~\ref{tab:llm_audit}).
Task~1 CUI alignment accuracy is $\geq$0.98 across all languages; Task~2 Tier-1 synonym accuracy is 1.00.
For Task~3 translations, ES achieves 1.00, FR 0.86, RU 0.92, while JA and ZH drop to 0.80 and 0.78 respectively, consistent with limited UMLS coverage for non-Latin scripts (Section~\ref{sec:limitations}).
On a 50-query French subset, an author independently evaluated translation fidelity under six predefined medical-translation criteria; inter-rater agreement with the LLM judge was 98\% (Cohen's $\kappa{=}0.92$; Gwet's AC1${=}0.97$), with the LLM judge identifying 7 of 8 human-flagged errors.

\begin{table}[!ht]
\centering
\caption{LLM-based quality audit accuracy per task and language ($n{=}50$ per cell). Evaluator: Claude Opus 4.7~(Anthropic, 2026).}
\label{tab:llm_audit}
\small
\begin{tabular}{@{}lcccccc@{}}
\toprule
\textbf{Task} & \textbf{EN} & \textbf{ES} & \textbf{FR} & \textbf{JA} & \textbf{ZH} & \textbf{RU} \\
\midrule
T1 (CUI align.) & .98 & 1.00 & --- & .98 & 1.00 & .98 \\
T2 (Tier-1 syn.) & 1.00 & --- & --- & --- & 1.00 & --- \\
T3 (Translation) & --- & 1.00 & .86 & .80 & .78 & .92 \\
\bottomrule
\end{tabular}
\end{table}

\section{Release and Societal Considerations}
\label{sec:release}

Evaluation scripts, benchmark splits, baseline results, and construction code are provided as supplementary material and will be hosted on HuggingFace Datasets.
Reproducing from raw sources requires UMLS 2025AB (NLM license) and BioASQ~13b (registration); the released package includes all derived artifacts needed for evaluation.
All data derives from licensed sources with no patient-identifiable information.
The fairness gap metric operationalizes cross-lingual equity; benchmark scores measure system capability, not clinical safety.

\section{Conclusion}

In this work, we introduce MMed-Bench-IR, a heterogeneous benchmark that jointly evaluates multilingual alignment and biomedical specialization in medical information retrieval across 6~languages and 3~writing systems.
Its three structurally distinct tasks share zero query and concept overlap, ensuring that aggregate scores reflect genuine capability breadth rather than mastery of a single skill.
Evaluation of ten systems across six paradigm families reveals consistent paradigm-level rankings, identifies concept discrimination as the hardest lexical bottleneck, and exposes the collapse of biomedical models on non-Latin scripts, failure modes detectable only through multi-axis evaluation.
We hope that MMed-Bench-IR serves as a foundation for developing retrieval systems that work equitably across languages in clinical settings.

\section*{Limitations}
\label{sec:limitations}

MMed-Bench-IR provides the first joint evaluation of multilingual alignment and biomedical specialization in medical retrieval, but several design choices constrain the current version.

Task~3 results for Japanese and Chinese should be interpreted as conservative estimates.
An LLM-based audit (Table~\ref{tab:llm_audit}) confirms that 78 to 80\% of JA/ZH translations preserve clinical meaning, while ES achieves 100\% and FR achieves 86\%, consistent with limited UMLS coverage for non-Latin scripts.
Future versions could benefit from ensemble translation or human post-editing for CJK languages.

The Task~2 corpus of 6{,}340 terms constrains negative discrimination difficulty, particularly at Tier~3 where only 325 confusion sets are available.
This limits the statistical power of Tier~3 evaluation, as reflected in the wide bootstrap confidence intervals (Table~\ref{tab:tiers}).
Scaling the confusion set corpus with additional UMLS source vocabularies is a priority for future versions.

Six languages exclude Arabic, Hindi, and other high-population languages with significant healthcare needs, though the construction framework is designed for community extension.
Task~1 excludes French because fewer than 50 French queries passed the CUI-tagging pipeline at the 500-concept granularity.
French is retained in Task~3 via NLLB-derived translations.

MMed-Embed is the only within-distribution system evaluated, and the current baselines do not include recent LLM-based embedding models (e.g., GritLM, E5-Mistral).
Community submissions would strengthen evaluation and broaden paradigm coverage beyond the six families tested here.

All three tasks rely on automated relevance judgments rather than expert human annotation.
While Section~\ref{sec:validity} demonstrates high agreement between LLM-based audits and human evaluation ($\kappa{=}0.92$), systematic biases may still exist for underrepresented languages where reference resources are sparse.
Incorporating expert clinician review across all six languages remains an important direction for future iterations.


\bibliography{ref}

@inproceedings{acharya-etal-2025-m3retrieve,
    title = "{M}3{R}etrieve: Benchmarking Multimodal Retrieval for Medicine",
    author = "Acharya, Arkadeep  and
      Ghosh, Akash  and
      Verma, Pradeepika  and
      Pasupa, Kitsuchart  and
      Saha, Sriparna  and
      Singh, Dr Priti",
    editor = "Christodoulopoulos, Christos  and
      Chakraborty, Tanmoy  and
      Rose, Carolyn  and
      Peng, Violet",
    booktitle = "Proceedings of the 2025 Conference on Empirical Methods in Natural Language Processing",
    month = nov,
    year = "2025",
    address = "Suzhou, China",
    publisher = "Association for Computational Linguistics",
    url = "https://aclanthology.org/2025.emnlp-main.771/",
    doi = "10.18653/v1/2025.emnlp-main.771",
    pages = "15263--15276",
    ISBN = "979-8-89176-332-6",
}

@article{anyaegbuna2026artificial,
  title={Artificial intelligence translation in healthcare: an urgent call for evidence-informed policy frameworks},
  author={Anyaegbuna, Chukwuebuka and Steele, Natasha and Liang, April Shichu and Ma, Stephen P and Lopez, Ivan and Chilukuri, Nymisha and Patel, Kavita and Schulman, Kevin and Chen, Jonathan H},
  journal={BMJ Health \& Care Informatics},
  volume={33},
  number={1},
  pages={e102007},
  year={2026}
}

@inproceedings{athar2025cure,
  title={CURE: A Dataset for Clinical Understanding \& Retrieval Evaluation},
  author={Athar Sheikh, Nadia and Buades Marcos, Daniel and Jousse, Anne-Laure and Oladipo, Akintunde and Rousseau, Olivier and Lin, Jimmy},
  booktitle={Proceedings of the 31st ACM SIGKDD Conference on Knowledge Discovery and Data Mining V. 2},
  pages={5270--5277},
  year={2025}
}

@article{bodenreider2004unified,
  title={The unified medical language system (UMLS): integrating biomedical terminology},
  author={Bodenreider, Olivier},
  journal={Nucleic acids research},
  volume={32},
  number={suppl\_1},
  pages={D267--D270},
  year={2004},
  publisher={Oxford University Press}
}

@inproceedings{chen-etal-2024-m3,
    title = "{M}3-Embedding: Multi-Linguality, Multi-Functionality, Multi-Granularity Text Embeddings Through Self-Knowledge Distillation",
    author = "Chen, Jianlyu  and
      Xiao, Shitao  and
      Zhang, Peitian  and
      Luo, Kun  and
      Lian, Defu  and
      Liu, Zheng",
    editor = "Ku, Lun-Wei  and
      Martins, Andre  and
      Srikumar, Vivek",
    booktitle = "Findings of the Association for Computational Linguistics: ACL 2024",
    month = aug,
    year = "2024",
    address = "Bangkok, Thailand",
    publisher = "Association for Computational Linguistics",
    url = "https://aclanthology.org/2024.findings-acl.137/",
    doi = "10.18653/v1/2024.findings-acl.137",
    pages = "2318--2335",
}

@inproceedings{conneau2020unsupervised,
  title={Unsupervised cross-lingual representation learning at scale},
  author={Conneau, Alexis and Khandelwal, Kartikay and Goyal, Naman and Chaudhary, Vishrav and Wenzek, Guillaume and Guzm{\'a}n, Francisco and Grave, Edouard and Ott, Myle and Zettlemoyer, Luke and Stoyanov, Veselin},
  booktitle={Proceedings of the 58th annual meeting of the association for computational linguistics},
  pages={8440--8451},
  year={2020}
}

@article{costa2022no,
  title={No language left behind: Scaling human-centered machine translation},
  author={Costa-Juss{\`a}, Marta R and Cross, James and {\c{C}}elebi, Onur and Elbayad, Maha and Heafield, Kenneth and Heffernan, Kevin and Kalbassi, Elahe and Lam, Janice and Licht, Daniel and Maillard, Jean and others},
  journal={arXiv preprint arXiv:2207.04672},
  year={2022}
}

@article{jin2023medcpt,
  title={Medcpt: Contrastive pre-trained transformers with large-scale pubmed search logs for zero-shot biomedical information retrieval},
  author={Jin, Qiao and Kim, Won and Chen, Qingyu and Comeau, Donald C and Yeganova, Lana and Wilbur, W John and Lu, Zhiyong},
  journal={Bioinformatics},
  volume={39},
  number={11},
  pages={btad651},
  year={2023},
  publisher={Oxford University Press}
}

@article{joo2023association,
  title={Association of language barriers with perioperative and surgical outcomes: a systematic review},
  author={Joo, Hyundeok and Fernandez, Alicia and Wick, Elizabeth C and Moreno Lepe, Gala and Manuel, Solmaz P},
  journal={JAMA Network Open},
  volume={6},
  number={7},
  pages={e2322743},
  year={2023}
}

@inproceedings{liu2021self,
  title={Self-alignment pretraining for biomedical entity representations},
  author={Liu, Fangyu and Shareghi, Ehsan and Meng, Zaiqiao and Basaldella, Marco and Collier, Nigel},
  booktitle={Proceedings of the 2021 conference of the North American chapter of the association for computational linguistics: human language technologies},
  pages={4228--4238},
  year={2021}
}

@inproceedings{louis2025colbert,
  title={Colbert-xm: A modular multi-vector representation model for zero-shot multilingual information retrieval},
  author={Louis, Antoine and Saxena, Vageesh Kumar and Van Dijck, Gijs and Spanakis, Gerasimos},
  booktitle={Proceedings of the 31st International Conference on Computational Linguistics},
  pages={4370--4383},
  year={2025}
}

@inproceedings{muennighoff2023mteb,
  title={Mteb: Massive text embedding benchmark},
  author={Muennighoff, Niklas and Tazi, Nouamane and Magne, Lo{\"\i}c and Reimers, Nils},
  booktitle={Proceedings of the 17th Conference of the European Chapter of the Association for Computational Linguistics},
  pages={2014--2037},
  year={2023}
}

@article{ortega2025language,
  title={Language equity in health technology for patients with Non--English language preference},
  author={Ortega, Pilar and Miller De Rutt{\'e}, Alyssia and Vela, M{\'o}nica},
  journal={JAMA Network Open},
  volume={8},
  number={2},
  pages={e2457424},
  year={2025}
}

@inproceedings{pfeiffer-etal-2022-lifting,
    title = "Lifting the Curse of Multilinguality by Pre-training Modular Transformers",
    author = "Pfeiffer, Jonas  and
      Goyal, Naman  and
      Lin, Xi Victoria  and
      Li, Xian  and
      Cross, James  and
      Riedel, Sebastian  and
      Artetxe, Mikel",
    editor = "Carpuat, Marine  and
      de Marneffe, Marie-Catherine  and
      Meza Ruiz, Ivan Vladimir",
    booktitle = "Proceedings of the 2022 Conference of the North American Chapter of the Association for Computational Linguistics: Human Language Technologies",
    month = jul,
    year = "2022",
    address = "Seattle, United States",
    publisher = "Association for Computational Linguistics",
    url = "https://aclanthology.org/2022.naacl-main.255/",
    doi = "10.18653/v1/2022.naacl-main.255",
    pages = "3479--3495",
}

@article{qiu2024towards,
  title={Towards building multilingual language model for medicine},
  author={Qiu, Pengcheng and Wu, Chaoyi and Zhang, Xiaoman and Lin, Weixiong and Wang, Haicheng and Zhang, Ya and Wang, Yanfeng and Xie, Weidi},
  journal={Nature Communications},
  volume={15},
  number={1},
  pages={8384},
  year={2024},
  publisher={Nature Publishing Group UK London}
}

@article{remy2023biolord,
  title={Biolord-2023: Semantic textual representations fusing llm and clinical knowledge graph insights},
  author={Remy, Fran{\c{c}}ois and Demuynck, Kris and Demeester, Thomas},
  journal={arXiv preprint arXiv:2311.16075},
  year={2023}
}

@book{robertson2009probabilistic,
  title={The probabilistic relevance framework: BM25 and beyond},
  author={Robertson, Stephen and Zaragoza, Hugo},
  volume={4},
  year={2009},
  publisher={Now Publishers Inc}
}

@article{sahni2023artificial,
  title={Artificial intelligence in US health care delivery},
  author={Sahni, Nikhil R and Carrus, Brandon},
  journal={New England Journal of Medicine},
  volume={389},
  number={4},
  pages={348--358},
  year={2023},
  publisher={Mass Medical Soc}
}

@inproceedings{
    thakur2021beir,
    title={{BEIR}: A Heterogeneous Benchmark for Zero-shot Evaluation of Information Retrieval Models},
    author={Nandan Thakur and Nils Reimers and Andreas R{\"u}ckl{\'e} and Abhishek Srivastava and Iryna Gurevych},
    booktitle={Thirty-fifth Conference on Neural Information Processing Systems Datasets and Benchmarks Track (Round 2)},
    year={2021},
    url={https://openreview.net/forum?id=wCu6T5xFjeJ}
}

@article{tsatsaronis2015overview,
  title={An overview of the BIOASQ large-scale biomedical semantic indexing and question answering competition},
  author={Tsatsaronis, George and Balikas, Georgios and Malakasiotis, Prodromos and Partalas, Ioannis and Zschunke, Matthias and Alvers, Michael R and Weissenborn, Dirk and Krithara, Anastasia and Petridis, Sergios and Polychronopoulos, Dimitris and others},
  journal={BMC bioinformatics},
  volume={16},
  number={1},
  pages={138},
  year={2015},
  publisher={Springer}
}

@article{volk2002semantic,
  title={Semantic annotation for concept-based cross-language medical information retrieval},
  author={Volk, Martin and Ripplinger, B{\"a}rbel and Vintar, {\v{S}}pela and Buitelaar, Paul and Raileanu, Diana and Sacaleanu, Bogdan},
  journal={International Journal of Medical Informatics},
  volume={67},
  number={1-3},
  pages={97--112},
  year={2002},
  publisher={Elsevier}
}

@article{wang2024apollo,
  title={Apollo: A lightweight multilingual medical LLM towards democratizing medical AI to 6B people},
  author={Wang, Xidong and Chen, Nuo and Chen, Junyin and Wang, Yidong and Zhen, Guorui and Zhang, Chunxian and Wu, Xiangbo and Hu, Yan and Gao, Anningzhe and Wan, Xiang and others},
  journal={arXiv preprint arXiv:2403.03640},
  year={2024}
}

@article{wang2024multilingual,
  title={Multilingual e5 text embeddings: A technical report},
  author={Wang, Liang and Yang, Nan and Huang, Xiaolong and Yang, Linjun and Majumder, Rangan and Wei, Furu},
  journal={arXiv preprint arXiv:2402.05672},
  year={2024}
}

@article{xiong2020approximate,
  title={Approximate nearest neighbor negative contrastive learning for dense text retrieval},
  author={Xiong, Lee and Xiong, Chenyan and Li, Ye and Tang, Kwok-Fung and Liu, Jialin and Bennett, Paul and Ahmed, Junaid and Overwijk, Arnold},
  journal={arXiv preprint arXiv:2007.00808},
  year={2020}
}

@article{yuan2022coder,
  title={CODER: Knowledge-infused cross-lingual medical term embedding for term normalization},
  author={Yuan, Zheng and Zhao, Zhengyun and Sun, Haixia and Li, Jiao and Wang, Fei and Yu, Sheng},
  journal={Journal of biomedical informatics},
  volume={126},
  pages={103983},
  year={2022},
  publisher={Elsevier}
}

@inproceedings{zhang2021mr,
  title={Mr. TyDi: A multi-lingual benchmark for dense retrieval},
  author={Zhang, Xinyu and Ma, Xueguang and Shi, Peng and Lin, Jimmy},
  booktitle={Proceedings of the 1st workshop on multilingual representation learning},
  pages={127--137},
  year={2021}
}

@article{zhang2023miracl,
  title={Miracl: A multilingual retrieval dataset covering 18 diverse languages},
  author={Zhang, Xinyu and Thakur, Nandan and Ogundepo, Odunayo and Kamalloo, Ehsan and Alfonso-Hermelo, David and Li, Xiaoguang and Liu, Qun and Rezagholizadeh, Mehdi and Lin, Jimmy},
  journal={Transactions of the Association for Computational Linguistics},
  volume={11},
  pages={1114--1131},
  year={2023},
  publisher={MIT Press One Broadway, 12th Floor, Cambridge, Massachusetts 02142, USA~…}
}

\appendix
\clearpage
\renewcommand{\thefigure}{A\arabic{figure}}
\renewcommand{\thetable}{A\arabic{table}}
\setcounter{figure}{0}
\setcounter{table}{0}

\section{Benchmark Construction Details}
\label{app:construction}

\subsection{Task~1: Data Statistics}

\begin{table}[!htbp]
\centering
\caption{Task~1 query distribution by language.}
\small
\begin{tabular}{@{}lccccc@{}}
\toprule
& \textbf{EN} & \textbf{ES} & \textbf{JA} & \textbf{RU} & \textbf{ZH} \\
\midrule
Queries & 3{,}696 & 1{,}693 & 346 & 303 & 89 \\
\bottomrule
\end{tabular}
\end{table}

Total corpus: 2{,}552 documents (1{,}290 positive + 1{,}262 hard negatives). All queries grounded with UMLS CUI tags (100\% coverage).

\subsection{Task~2: Multi-Encoder Validation}

Three independent encoders (BGE-M3, E5-large, SapBERT) compute pairwise cosine similarities for each confusion set.
A set is accepted only if $\geq$2/3~encoders agree on the tier assignment based on cosine similarity thresholds:
Tier~1: $\geq$0.70; Tier~2: 0.40--0.75; Tier~3: 0.15--0.45.

\begin{table}[!htbp]
\centering
\caption{Tier validation statistics. Mean cosine similarity (3-encoder average) confirms monotonic tier ordering with non-overlapping confidence intervals.}
\footnotesize
\begin{tabular}{@{}lcccc@{}}
\toprule
\textbf{Tier} & \textbf{Mean$\pm$Std} & \textbf{Min} & \textbf{Max} & \textbf{N} \\
\midrule
T1 (Synonym) & $.790\pm.089$ & .525 & 1.000 & 3{,}005 \\
T2 (Sibling) & $.655\pm.047$ & .493 & .803 & 1{,}645 \\
T3 (Related) & $.514\pm.029$ & .433 & .577 & 325 \\
\midrule
Rejected & --- & --- & --- & 967 \\
Re-tiered & --- & --- & --- & 1{,}597 \\
\bottomrule
\end{tabular}
\end{table}

\subsection{Task~2: Leave-One-Validator-Out Analysis}

To assess whether tier assignments depend on any single validator, we re-run the validation using each 2-encoder pair (requiring 2/2 agreement) and compare with the full 3-encoder result.

\begin{table}[!htbp]
\centering
\caption{Leave-one-validator-out sensitivity. Among sets accepted by each 2-encoder pair, we report the fraction matching the full 3-encoder tier assignment. No set shifts to a non-adjacent tier in any condition.}
\small
\begin{tabular}{@{}lccc@{}}
\toprule
\textbf{Excluded} & \textbf{Accepted} & \textbf{Same tier} & \textbf{Stability} \\
\midrule
BGE-M3 & 1{,}212 & 1{,}212 & 100.0\% \\
E5-large & 3{,}477 & 2{,}968 & 85.4\% \\
SapBERT & 2{,}879 & 2{,}879 & 100.0\% \\
\bottomrule
\end{tabular}
\end{table}

The stricter 2/2 agreement threshold naturally reduces the number of accepted sets, but among those accepted, tier assignments are highly stable.
When E5-large is excluded, 509 sets (14.6\%) shift between adjacent tiers (predominantly Tier~1 $\to$ Tier~2), reflecting differences in the encoders' similarity calibration rather than fundamental tier ambiguity.

\subsection{Task~3: Translation Quality Assurance}

\paragraph{Concept fidelity.}
A bilingual medical lexicon built from 587K UMLS English terms mapped to target-language equivalents verifies that medical concepts are preserved in translation.
Standard local-language medical terminology is accepted, not just English surface forms.

\paragraph{Back-translation consistency.}
Each translation is back-translated to English via NLLB-200 and compared with the original via token-overlap F1.

\paragraph{Audit status thresholds.}
\textbf{Pass}: term preservation rate $\geq$0.9 AND back-translation F1 $\geq$0.8.
\textbf{Fail}: term preservation $<$0.7 OR back-translation F1 $<$0.6.
\textbf{Flag}: all other cases (at least one criterion met but not all).
Queries with status ``pass'' or ``flag'' form the official evaluation subset.

\paragraph{CUI tagging thresholds.}
Task~1 queries are CUI-tagged via a cascading pipeline:
(1)~exact lexical match against UMLS preferred terms (confidence $\geq$0.9, substring match requires $\geq$3 characters);
(2)~biomedical linker fallback (confidence $\geq$0.7);
(3)~null if both fail.

\begin{table}[!htbp]
\centering
\caption{Translation QA results by language. This table covers the 5 translated languages only (English source excluded). Official subset includes \texttt{pass} + \texttt{flag} queries. Total official = 1{,}368 translated + 340 English source = 1{,}708.}
\small
\begin{tabular}{@{}lccccc@{}}
\toprule
\textbf{Lang} & \textbf{Pass} & \textbf{Flag} & \textbf{Fail} & \textbf{Official} & \textbf{Total} \\
\midrule
ES & 82 & 249 & 9 & 331 & 340 \\
FR & 53 & 271 & 16 & 324 & 340 \\
JA & 2 & 213 & 125 & 215 & 340 \\
ZH & 9 & 209 & 122 & 218 & 340 \\
RU & 6 & 274 & 60 & 280 & 340 \\
\midrule
Total & 152 & 1{,}216 & 332 & 1{,}368 & 1{,}700 \\
\bottomrule
\end{tabular}
\end{table}

\subsection{Train/Test Separation Verification}

\begin{table}[!htbp]
\centering
\caption{Query text overlap between MMed-Embed training data (137K rows) and benchmark test sets.}
\small
\begin{tabular}{@{}lcc@{}}
\toprule
\textbf{Task} & \textbf{Overlap} & \textbf{Rate} \\
\midrule
Task~1 (6{,}127 queries) & 2 & 0.03\% \\
Task~2 (4{,}143 queries) & 20 & 0.48\% \\
Task~3 (2{,}040 queries) & 0 & 0.00\% \\
\bottomrule
\end{tabular}
\end{table}

\begin{table*}[!htbp]
\centering
\caption{MMed-Embed 2$\times$2 factorial ablation with bootstrap 95\% CIs. All variants: 137K rows, 10 epochs, identical hyperparameters.}
\label{tab:ablation}
\small
\begin{tabular}{@{}lcccccc@{}}
\toprule
& \textbf{Loss} & \textbf{Negatives} & \textbf{Task 1} & \textbf{Task 2} & \textbf{Task 3} & \textbf{MMed-IR} \\
\midrule
A0 & --- & --- & .344\,{\scriptsize$\pm$.007} & .432\,{\scriptsize$\pm$.010} & .264\,{\scriptsize$\pm$.013} & .347 \\
A1 & $\mathcal{L}_\text{ret}$ & random & \textbf{.388}\,{\scriptsize$\pm$.007} & .454\,{\scriptsize$\pm$.011} & \textbf{.267}\,{\scriptsize$\pm$.013} & \textbf{.369} \\
A2 & $\mathcal{L}_\text{ret}{+}\mathcal{L}_\text{topo}$ & random & .387\,{\scriptsize$\pm$.007} & .451\,{\scriptsize$\pm$.011} & .265\,{\scriptsize$\pm$.013} & .368 \\
A3 & $\mathcal{L}_\text{ret}$ & ANCE & .370\,{\scriptsize$\pm$.007} & \textbf{.459}\,{\scriptsize$\pm$.011} & .265\,{\scriptsize$\pm$.013} & .364 \\
A4 & $\mathcal{L}_\text{ret}{+}\mathcal{L}_\text{topo}$ & ANCE & .370\,{\scriptsize$\pm$.007} & \textbf{.459}\,{\scriptsize$\pm$.011} & .263\,{\scriptsize$\pm$.013} & .364 \\
\bottomrule
\end{tabular}
\end{table*}

Task~2 overlaps are short UMLS terms (e.g., ``ATC'', ``airway resistance'') common to both training vocabulary and confusion set members. No semantic leakage is present.

\section{Reference Model: MMed-Embed}
\label{app:reference}

Training objective, two-stage procedure, and key hyperparameters are described in Section~\ref{sec:reference}.

\paragraph{Hard negative mining.}
Hard negatives are mined using ANCE-style~\citep{xiong2020approximate} retrieval from the Stage-1 model's embeddings: for each query, the top-$k$ most similar non-positive documents are selected ($k{=}15$).
To avoid false negatives, candidates with similarity $\geq$95\% of the positive score are filtered out (positive-aware filtering).

\paragraph{Ablation study.}
Table~\ref{tab:ablation} presents a 2$\times$2 factorial ablation isolating the contributions of topology-aware loss ($\mathcal{L}_\text{topo}$) and ANCE-style hard negative mining.
All variants use identical hyperparameters: 137K training rows, 10~epochs, effective batch 1{,}024, lr $2{\times}10^{-6}$, starting from BGE-M3.

The dominant effect is fine-tuning itself: Stage-1 contrastive training (A1) improves MMed-IR by +0.022 over the BGE-M3 base.
Neither $\mathcal{L}_\text{topo}$ nor hard negative mining produces a statistically significant additional gain under controlled conditions:
A1 vs.\ A2 ($+\mathcal{L}_\text{topo}$, random neg): $\Delta$MMed-IR $= -$0.001;
A1 vs.\ A3 (mined neg, $\mathcal{L}_\text{ret}$ only): $\Delta$MMed-IR $= -$0.005.
Hard negative mining shifts the performance profile, improving Task~2 (+0.005) at the cost of Task~1 ($-$0.018), but does not improve the aggregate.
This suggests that for MMed-Bench-IR, the primary benefit comes from domain-specific contrastive fine-tuning on multilingual medical data, rather than from loss function design or negative mining strategy.
The marginal gain of fine-tuning over zero-shot BGE-M3 ($+$0.022 MMed-IR) further indicates that the benchmark's core challenge is architecture-level multilingual alignment, not training data availability.

Full hyperparameters are in Table~\ref{tab:hyperparams}.

\begin{table*}[!htbp]
\centering
\caption{MMed-Embed training hyperparameters.}
\label{tab:hyperparams}
\small
\begin{tabular}{@{}ll@{}}
\toprule
\textbf{Parameter} & \textbf{Value} \\
\midrule
Base model & BAAI/bge-m3 (568M params) \\
Pooling & Mean pooling + L2 normalization \\
Training data format & FlagEmbedding (\texttt{query}, \texttt{pos}, \texttt{neg}) \\
Optimizer & AdamW (weight decay 0.01) \\
LR schedule & Linear warmup (10\%) + linear decay \\
Temperature $\tau$ & 0.02 \\
Max query length & 256 tokens \\
Max passage length & 256 tokens \\
Hard negatives per query & 15 random \\
In-batch negatives & Cross-GPU gathered (effective $B{-}1$) \\
Gradient checkpointing & Enabled \\
Precision & Mixed (bf16) \\
Hardware & 8$\times$NVIDIA B200 GPUs \\
DDP backend & NCCL \\
Seed & 42 \\
\midrule
Training data & 137K rows (MMedBench 40K + UMLS 87K + xBioASQ 10K) \\
Negatives & 15 random per query \\
Loss & $\mathcal{L}_\text{ret}$ (contrastive) \\
Per-device batch size & 128 (effective 1{,}024) \\
Learning rate & $2{\times}10^{-6}$ \\
Epochs & 10 (1{,}330 steps) \\
\bottomrule
\end{tabular}
\end{table*}

\section{Full Per-Model Results}
\label{app:full_results}

All tables below include all ten evaluated systems across six paradigm families.

\paragraph{Statistical framework.}
All 95\% CIs reported in this paper use query-level bootstrap resampling ($n{=}2{,}000$ iterations) applied uniformly across all tables and headline claims.
For each metric (R@1, R@5, nDCG@10), per-query scores are resampled with replacement and the 2.5th/97.5th percentiles form the CI.
For composite metrics (Task~1 score = mean of per-query nDCG@10 and R@1; Task~2 score = mean of three tier R@1s; MMed-IR = mean of three task scores), we resample per-query scores at each level and propagate through the composition.
Fairness gap CIs are computed by bootstrapping the max$-$min of per-language means, capturing the full uncertainty in both the maximum and minimum languages.
All CIs are reported as $\pm$ half-width of the 95\% interval.

\subsection{Task~1: Per-Language nDCG@10}

Table~\ref{tab:app_task1_lang} reports per-language nDCG@10 for all ten systems.
Biomedical-only models show near-zero scores on Japanese, while multilingual models maintain moderate performance across all languages.

\begin{table*}[!htbp]
\centering
\caption{Task~1 per-language nDCG@10 with bootstrap 95\% CIs for all ten systems.}
\label{tab:app_task1_lang}
\small
\begin{tabular}{@{}llccccc@{}}
\toprule
\textbf{Model} & \textbf{Par.} & \textbf{EN} & \textbf{ES} & \textbf{JA} & \textbf{RU} & \textbf{ZH} \\
\midrule
BM25 & Lex & .162\,{\scriptsize$\pm$.009} & .133\,{\scriptsize$\pm$.012} & .004\,{\scriptsize$\pm$.005} & .064\,{\scriptsize$\pm$.018} & .077\,{\scriptsize$\pm$.037} \\
BioLORD & Bio & .367\,{\scriptsize$\pm$.009} & .292\,{\scriptsize$\pm$.015} & .000\,{\scriptsize$\pm$.000} & .058\,{\scriptsize$\pm$.020} & .032\,{\scriptsize$\pm$.027} \\
SapBERT & Bio & .428\,{\scriptsize$\pm$.009} & .315\,{\scriptsize$\pm$.015} & .001\,{\scriptsize$\pm$.002} & .091\,{\scriptsize$\pm$.024} & .011\,{\scriptsize$\pm$.012} \\
Hybrid & Hyb & .323\,{\scriptsize$\pm$.009} & .250\,{\scriptsize$\pm$.013} & .185\,{\scriptsize$\pm$.024} & .276\,{\scriptsize$\pm$.028} & .145\,{\scriptsize$\pm$.038} \\
BGE-M3 & Multi & .590\,{\scriptsize$\pm$.010} & .493\,{\scriptsize$\pm$.014} & .254\,{\scriptsize$\pm$.033} & .407\,{\scriptsize$\pm$.032} & .148\,{\scriptsize$\pm$.048} \\
E5-large & Multi & .570\,{\scriptsize$\pm$.009} & .456\,{\scriptsize$\pm$.015} & .248\,{\scriptsize$\pm$.029} & .332\,{\scriptsize$\pm$.028} & .061\,{\scriptsize$\pm$.037} \\
ColBERT-XM & Late-int & .516\,{\scriptsize$\pm$.010} & .304\,{\scriptsize$\pm$.014} & .151\,{\scriptsize$\pm$.024} & .262\,{\scriptsize$\pm$.030} & .057\,{\scriptsize$\pm$.036} \\
Reranker & 2-stg & .628\,{\scriptsize$\pm$.009} & .566\,{\scriptsize$\pm$.014} & .316\,{\scriptsize$\pm$.034} & .482\,{\scriptsize$\pm$.031} & .269\,{\scriptsize$\pm$.051} \\
MMed-Embed$^\dagger$ & M+Med & .638\,{\scriptsize$\pm$.009} & .593\,{\scriptsize$\pm$.011} & .347\,{\scriptsize$\pm$.033} & .539\,{\scriptsize$\pm$.027} & .385\,{\scriptsize$\pm$.053} \\
MMed+Rer.$^\dagger$ & M+R & \textbf{.645}\,{\scriptsize$\pm$.009} & \textbf{.607}\,{\scriptsize$\pm$.013} & \textbf{.392}\,{\scriptsize$\pm$.035} & \textbf{.537}\,{\scriptsize$\pm$.028} & \textbf{.467}\,{\scriptsize$\pm$.049} \\
\bottomrule
\end{tabular}
\end{table*}

\subsection{Task~2: Per-Tier Recall@5}

Table~\ref{tab:app_task2_r5} extends the Recall@1 analysis in the main text to Recall@5.
The tier ordering (Synonym, Sibling, Related) is preserved across all models at this cutoff.

\begin{table*}[!htbp]
\centering
\caption{Task~2 Recall@5 by tier with bootstrap 95\% CIs for all ten systems.}
\label{tab:app_task2_r5}
\small
\begin{tabular}{@{}llccc@{}}
\toprule
\textbf{Model} & \textbf{Par.} & \textbf{Synonym} & \textbf{Sibling} & \textbf{Related} \\
\midrule
BM25 & Lex & .265\,{\scriptsize$\pm$.016} & .105\,{\scriptsize$\pm$.014} & .004\,{\scriptsize$\pm$.006} \\
BioLORD & Bio & .613\,{\scriptsize$\pm$.017} & .273\,{\scriptsize$\pm$.022} & .044\,{\scriptsize$\pm$.034} \\
SapBERT & Bio & .729\,{\scriptsize$\pm$.016} & .380\,{\scriptsize$\pm$.022} & .015\,{\scriptsize$\pm$.013} \\
Hybrid & Hyb & .829\,{\scriptsize$\pm$.013} & .439\,{\scriptsize$\pm$.023} & .025\,{\scriptsize$\pm$.019} \\
ColBERT-XM & Late-int & .921\,{\scriptsize$\pm$.010} & .494\,{\scriptsize$\pm$.022} & .023\,{\scriptsize$\pm$.018} \\
BGE-M3 & Multi & .982\,{\scriptsize$\pm$.004} & .631\,{\scriptsize$\pm$.021} & .033\,{\scriptsize$\pm$.024} \\
E5-large & Multi & .977\,{\scriptsize$\pm$.005} & .637\,{\scriptsize$\pm$.021} & .048\,{\scriptsize$\pm$.032} \\
Reranker & 2-stg & .985\,{\scriptsize$\pm$.004} & .676\,{\scriptsize$\pm$.020} & .044\,{\scriptsize$\pm$.027} \\
MMed-Embed$^\dagger$ & M+Med & .989\,{\scriptsize$\pm$.003} & .710\,{\scriptsize$\pm$.021} & .048\,{\scriptsize$\pm$.026} \\
MMed+Rer.$^\dagger$ & M+R & \textbf{.988}\,{\scriptsize$\pm$.004} & \textbf{.723}\,{\scriptsize$\pm$.020} & \textbf{.048}\,{\scriptsize$\pm$.028} \\
\bottomrule
\end{tabular}
\end{table*}

\subsection{Fairness Gaps}

Table~\ref{tab:app_fairness} reports cross-language fairness gaps across all three tasks.
MMed-Embed with Reranker achieves the lowest gap on all tasks, while biomedical-only models show the largest gaps, consistent with Finding~C.

\begin{table}[!htbp]
\centering
\caption{Cross-language fairness gap (max $-$ min nDCG@10 across languages) per task for all ten systems. Lower is more equitable.}
\label{tab:app_fairness}
\footnotesize
\begin{tabular}{@{}llccc@{}}
\toprule
\textbf{Model} & \textbf{Par.} & \textbf{T1} & \textbf{T2} & \textbf{T3} \\
\midrule
BM25 & Lex & .162 & .012 & .485 \\
BioLORD & Bio & .367 & .353 & .693 \\
SapBERT & Bio & .427 & .520 & .762 \\
BM25+BGE-M3 & Hyb & .189 & .104 & .400 \\
BGE-M3 & Multi & .441 & .308 & .235 \\
E5-large & Multi & .509 & .296 & .394 \\
ColBERT-XM & Late & .459 & .331 & .369 \\
BGE-M3+Rer. & 2-stg & .358 & .208 & .180 \\
MMed-Embed$^\dagger$ & M+M & .291 & .126 & .221 \\
MMed+Rer.$^\dagger$ & M+R & .253 & .104 & .170 \\
\bottomrule
\end{tabular}
\end{table}

\subsection{Task~3: Translation Sensitivity}

To verify that the QA filtering does not shape benchmark outcomes, we compare model rankings under three query subsets: pass-only (492~queries, Latin-script dominated), official pass+flag (1{,}708), and all translated (2{,}040).
This analysis uses five zero-shot baselines (excluding hybrid, reranker, and MMed-Embed) to isolate the effect of QA filtering on ranking order.

\begin{table}[!htbp]
\centering
\caption{Task~3 weighted nDCG@10 under different QA subsets. Model ranking is perfectly preserved between official and all-translated ($\rho{=}1.0$) and nearly so with pass-only ($\rho{=}0.9$).}
\small
\begin{tabular}{@{}lccc@{}}
\toprule
\textbf{Model} & \textbf{Pass-only} & \textbf{Official} & \textbf{All translated} \\
\midrule
BM25 & 0.387 & 0.172 & 0.154 \\
BioLORD & 0.638 & 0.344 & 0.304 \\
SapBERT & 0.698 & 0.383 & 0.339 \\
BGE-M3 & 0.843 & 0.784 & 0.769 \\
E5-large & 0.868 & 0.778 & 0.755 \\
\bottomrule
\end{tabular}
\end{table}

\section{Benchmark Card}
\label{app:card}

\paragraph{Task type.} Information retrieval (ranking).
\paragraph{Languages.} English, Spanish, French, Japanese, Chinese, Russian (6~languages, 3~writing systems).
\paragraph{Domains.} Biomedical (UMLS-grounded concepts, BioASQ questions).
\paragraph{Data sources.} UMLS 2025AB (NLM license required), BioASQ 13b (registration required).
\paragraph{Annotation method.} Automatic construction with algorithmic validation:
UMLS CUI grounding for Task~1; 3-encoder majority vote for Task~2; UMLS concept-fidelity lexicon + back-translation consistency for Task~3.
\paragraph{Known limitations.} See Section~\ref{sec:limitations}.
Non-Latin-script translation quality is bounded by UMLS coverage for those languages.
\paragraph{Maintenance plan.} Benchmark hosted on HuggingFace Datasets.
Human audit results and version updates documented in the dataset card.
\paragraph{Intended use.} Evaluation of multilingual medical retrieval systems.
Not intended as training data.
\paragraph{Prohibited use.} Clinical decision support without further validation.

\section{Qualitative Examples}
\label{app:qualitative}

Tables~\ref{tab:qual_task1}--\ref{tab:qual_task3} present representative examples from each task, drawn from actual benchmark data.
These examples illustrate the cross-lingual and concept-level challenges that MMed-Bench-IR is designed to evaluate.

\begin{table*}[t]
\centering
\caption{\textbf{Task~1 qualitative examples.}
Two cross-lingual QA retrieval examples spanning Latin, CJK, and Cyrillic scripts.
All positives share the same UMLS CUI as the English query.}
\label{tab:qual_task1}
\small
\begin{tabular}{@{}llp{10cm}l@{}}
\toprule
\textbf{Example} & \textbf{Lang} & \textbf{Text} & \textbf{Role} \\
\midrule
\multirow{4}{*}{\shortstack[l]{Atrial Tachycardia\\Ectopic\\(C0039234)}}
  & EN & ATRIAL TACHYCARDIA ECTOPIC & Query \\
  & ES & Latidos cardiacos anormalmente r\'{a}pidos que se originan de uno o m\'{a}s focos autom\'{a}ticos en el ATRIO CARD\'{I}ACO... & Positive \\
  & JA & \begin{CJK}{UTF8}{min}心房異所性頻拍\end{CJK} & Positive \\
  & RU & EKTOPICHESKAIA PREDSERDNAIA TAKHIKARDIIA & Positive \\
\midrule
\multirow{4}{*}{\shortstack[l]{Lip/oral cavity\\cancer\\(C0861542)}}
  & EN & Lip and/or oral cavity cancer stage unspecified & Query \\
  & ES & C\'{a}ncer del labio o de la cavidad oral, estadio no especificado & Positive \\
  & JA & \begin{CJK}{UTF8}{min}口唇および口腔内癌、病期不明\end{CJK} & Positive \\
  & RU & Rak guby i (ili) rotovoj polosti neutochnennoj stadii & Positive \\
\bottomrule
\end{tabular}
\end{table*}

\begin{table*}[t]
\centering
\caption{\textbf{Task~2 qualitative examples.}
Three difficulty tiers with increasing surface-form divergence.
Encoder cosine similarities (BGE-M3 / E5-large / SapBERT) are shown for Tiers~2 and~3.
Tier~3 exhibits near-zero lexical overlap across languages, rendering BM25 ineffective.}
\label{tab:qual_task2}
\small
\begin{tabular}{@{}llp{6cm}lp{4cm}@{}}
\toprule
\textbf{Tier} & \textbf{Lang} & \textbf{Text} & \textbf{CUI} & \textbf{Note} \\
\midrule
\multirow{4}{*}{\shortstack[l]{Tier~1\\(Synonym)\\Easy}}
  & EN & Drug-induced asthma & C0340067 & \multirow{4}{*}{\shortstack[l]{High cross-lingual\\cohesion; lexical\\overlap partially\\preserved}} \\
  & JA & \begin{CJK}{UTF8}{min}薬物誘発性喘息\end{CJK} & C0340067 & \\
  & RU & Astma, vyzvannaia lekarstvennym sredstvom & C0340067 & \\
  & ES & Asma inducida por f\'{a}rmacos & C0340067 & \\
\midrule
\multirow{4}{*}{\shortstack[l]{Tier~2\\(Sibling)\\Medium}}
  & EN & Bowel infarction & C0241950 & \multirow{4}{*}{\shortstack[l]{BGE-M3: 0.61\\E5-large: 0.88\\SapBERT: 0.43\\Moderate divergence}} \\
  & JA & \begin{CJK}{UTF8}{min}腸梗塞\end{CJK} & C0241950 & \\
  & RU & Infarkt kishechnika & C0241950 & \\
  & ES & Infarto intestinal & C0241950 & \\
\midrule
\multirow{4}{*}{\shortstack[l]{Tier~3\\(Related)\\Hard}}
  & EN & ADP Phosphocreatine Phosphotransferase & C0010287 & \multirow{4}{*}{\shortstack[l]{BGE-M3: 0.39\\E5-large: 0.83\\SapBERT: 0.43\\\textbf{BM25 fails entirely}}} \\
  & JA & CK\begin{CJK}{UTF8}{min}アイソザイム\end{CJK} & C0010287 & \\
  & RU & KREATINFOSFOKINAZA & C0010287 & \\
  & ES & Creatina Fosfoquinasa & C0010287 & \\
\bottomrule
\end{tabular}
\end{table*}

\begin{table*}[t]
\centering
\caption{\textbf{Task~3 qualitative example.}
EGFR mutation query translated to 6~languages via NLLB-200 with two-stage translation QA.
Audit status reflects translation quality gating: PASS (both checks satisfied), FLAG (one check satisfied).
Latin-script translations pass more reliably than CJK translations, consistent with limited UMLS coverage for non-Latin scripts.}
\label{tab:qual_task3}
\small
\begin{tabular}{@{}llp{9cm}l@{}}
\toprule
\textbf{Role} & \textbf{Lang} & \textbf{Text} & \textbf{Audit} \\
\midrule
\multicolumn{4}{@{}l}{\textit{Queries (QID: 67da0c0f)}} \\
Source & EN & What is the most common EGFR mutation in glioblastoma multiforme? & --- \\
Translation & ES & \textquestiondown Cu\'{a}l es la mutaci\'{o}n EGFR m\'{a}s com\'{u}n en el glioblastoma multiforme? & PASS \\
Translation & FR & Quelle est la mutation EGFR la plus fr\'{e}quente dans le glioblastome multiforme~? & FLAG \\
Translation & JA & \begin{CJK}{UTF8}{min}膠芽腫で最も一般的なEGFR変異は?\end{CJK} & FLAG \\
Translation & ZH & \begin{CJK}{UTF8}{min}什么是最常见的EGFR突变在多形质母细胞瘤?\end{CJK} & FLAG \\
Translation & RU & Kakova naibolee rasprostranennaia mutatsiia EGFR v mul'tiformennoj glioblastome? & FLAG \\
\midrule
\multicolumn{4}{@{}l}{\textit{Gold evidence snippets (BioASQ 13b)}} \\
Evidence & EN & The epidermal growth factor receptor variant type III (EGFRvIII) is the most common mutation of EGFR in glioblastoma multiforme (GBM) and is found in approximately 25\% of all GBMs. & --- \\
Evidence & EN & In this study we determined whether EGFR amplification and expression of the most common mutation in GBMs (EGFRvIII) is retained at tumor recurrence. & --- \\
\bottomrule
\end{tabular}
\end{table*}

\end{document}